# Variational Policy for Guiding Point Processes

**Yichen Wang** [1] **Grady Williams** [2] **Evangelos Theodorou** [2] **Le Song** [1]

## Abstract

Temporal point processes have been widely applied to model event sequence data generated by online users. In this paper, we consider the problem of how to design the optimal control policy for point processes, such that the stochastic system driven by the point process is steered to a target state. In particular, we exploit the key insight to view the stochastic optimal control problem from the perspective of optimal measure and variational inference. We further propose a convex optimization framework and an efficient algorithm to update the policy adaptively to the current system state. Experiments on synthetic and real-world data show that our algorithm can steer the user activities much more accurately and efficiently than other stochastic control methods.

## 1. Introduction

Nowadays, user generated event data are becoming increasingly available. Each user is typically logged in the database with the precise time-stamp of the event, together with additional context such as tag, text, image, and video. Furthermore, these data are generated in an asynchronous fashion since any user can generate an event at any time and there may not be any coordination or synchronization between two events. Among different representations of user behaviors, temporal point processes have been widely applied to model the complex dynamics of online user behaviors (Zhou et al., 2013; Du et al., 2015; Lian et al., 2014; 2015; He et al., 2015; 2016; Dai et al., 2016; Wang et al., 2016a;b;c).

In spite of the broad applicability of point processes, there is little work in the area of controlling these processes to influence user behaviors. In this paper, we study the problem of designing the best intervention policy to influence the intensity function of point processes, such that user behaviors can be influenced towards a target state.

[1]College of Computing, Georgia Tech, Atlanta, GA, USA [2]School of Aerospace Engineering, Georgia Tech. Correspondence to: Yichen Wang <yichen.wang@gatech.edu>.

 

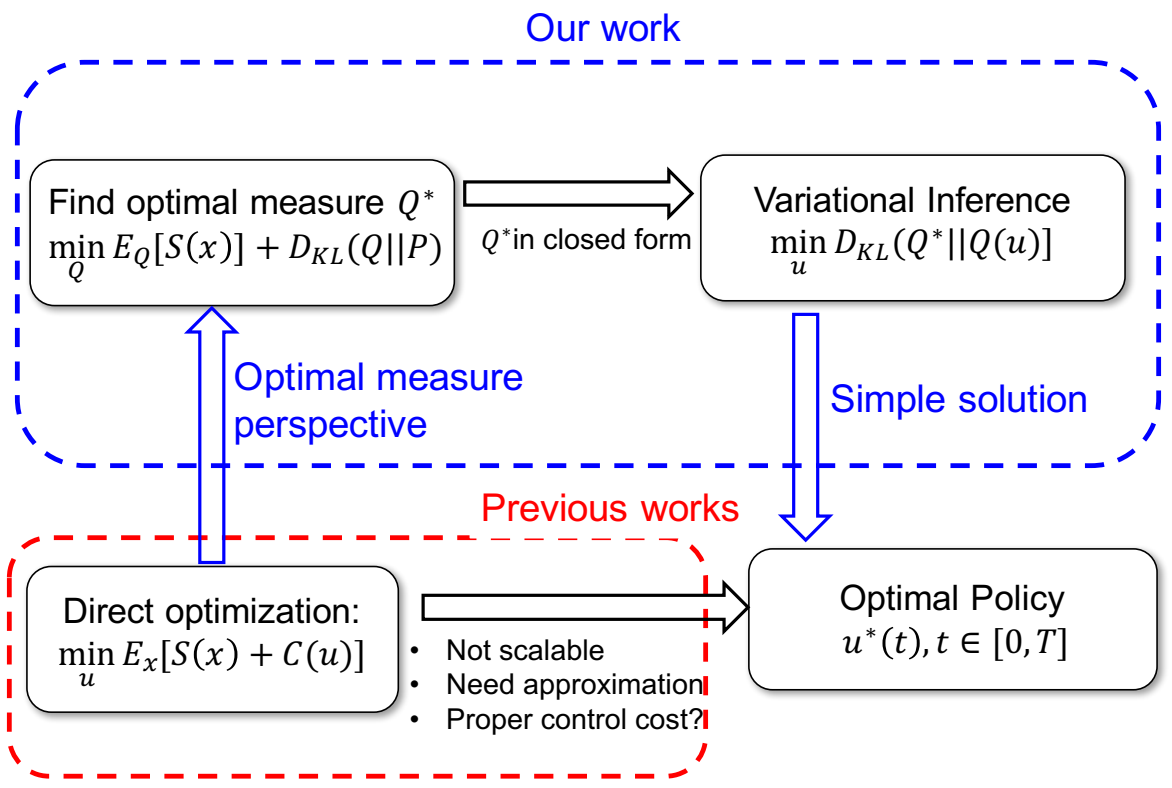

*Figure 1.* Illustration of the measure-theoretic view and benefit of our framework compared with existing approaches.

A framework for doing this is critically important. For example, government agents may want to effectively suppress the spread of terrorist propaganda, which is important for understanding the vulnerabilities of social networks and increasing their resilience to rumor and false information; online merchants may want to promote users' frequency of visiting the website to increase sales; administrators of Q&A sites such as StackOverflow design various badges to motivate users to answer questions and provide feedbacks to increase the online engagement (Anderson et al., 2013); to gain more attention, a broadcaster on Twitter may want to design a smart tweeting strategy such that his posts always remain on top of his followers' feeds (Karimi et al., 2016).

Interestingly, the social science setting also introduces new challenges. Previous stochastic optimal control methods (Boel & Varaiya, 1977; Pham, 1998; Oksendal & Sulem, 2005; Hanson, 2007) in robotics are not applicable for four reasons: (i) they mostly focus on the cases where the policy is in the drift part of the system, which is quite different from our case where the policy is on the intensity function; (ii) they require linear approximations of the nonlinear system and quadratic approximations of the objective function; (iii) to obtain a feedback control policy, these methods require the solution of the Hamilton-Jacobi-Bellman (HJB) Partial Differential Equation, which have severe limitations in scalability and feasibility to the nonlinear systems, especially in social applications where the system's dimension is huge; (iv) the systems they study are driven by Wiener processes and Poisson processes. However, social sciences require us to consider more advanced processes, such as Hawkes processes, which are models for long term memory process and mutual exciting phenomena in social interactions.

To address these limitations, we propose an efficient framework by exploiting the novel view of measure-theoretic formulation and variational inference. Figure 1 illustrates our method. We make the following contributions:

**Unified framework**. Our work offers a generic way to control nonlinear stochastic differential equations driven by point processes with stochastic intensities. Unlike prior works (Oksendal & Sulem, 2005), no approximations of the system or the objective function are needed.

**Natural control cost**. Our framework provides a meaningful control cost function to optimize: it arises naturally from the structure of the stochastic dynamics. This property is in stark contrast with the stochastic dynamic programming methods in control theory, where the control cost is imposed beforehand, despite the form of the dynamics.

**Superior performance**. We propose a scalable model predictive control algorithm. The control policy is computed with forward sampling; hence it is scalable with parallel sampling and runs in real time. Moreover, it enjoys superior empirical performance on diverse social applications.

## 2. Background and Preliminaries

**Point processes.** A temporal point process (Aalen et al., 2008) is a random process whose realization consists of a list of discrete events localized in time, $\{t_i\}$. It is widely applied to model user-generated event data and user behavior patterns (Farajtabar et al., 2014; 2015; Pan et al., 2016; Tan et al., 2016; Wang et al., 2016a;b;c; 2017).

The point process can also be represented as a counting process, $N(t)$, which records the number of events before time $t$. An important way to characterize it is via the conditional intensity function $\lambda(t)$ — a stochastic model for the time of the next event given historical events, $\mathcal{H}(t) = \{t_i | t_i < t\}$. It is the probability of observing a new event on $[t, t+\mathrm{d}t)$, *i.e.*,

$$\lambda(t)\mathrm{d}t := \mathbb{P}\left\{\text{event in } [t, t+\mathrm{d}t) | \mathcal{H}(t)\right\} = \mathbb{E}[\mathrm{d}N(t)|\mathcal{H}(t)]$$

where one typically assumes that only one event happens in a small window of size $\mathrm{d}t$, *i.e.*, $\mathrm{d}N(t) \in \{0, 1\}$.

The function form of the intensity is often designed to capture the phenomena of interests. Some useful forms include: *(i) Poisson process*: the intensity is independent of history; *(ii) Hawkes process* (Hawkes, 1971): It models the mutual excitation between events, and the intensity of a user $i$ depends on events from a collection of $M$ users:

$$\lambda_i(t) = \mu_i + \sum\nolimits_{j=1}^{M} \alpha_{ij} \sum\nolimits_{t_j \in \mathcal{H}_j(t)} \kappa_\omega(t - t_j), \qquad (1)$$

where $\kappa_\omega(t) = \exp(-\omega t)$ is a triggering kernel that models the decay of past events' influence, $\mu_i \geqslant 0$ is the base intensity, $N_j(t)$ is the point process representing the historical events $\mathcal{H}_j(t)$ from user $j$, and $\alpha_{ij} \geqslant 0$ models the strength of influence from user $j$ to user $i$. Here, the occurrence of each historical event increases the intensity by a certain amount determined by $\kappa_\omega(t)$ and the weight $\alpha_{ij}$, making $\lambda_i(t)$ history dependent and a stochastic process by itself.

The key rationale of using point processes for user behaviors is that these models treat time as a continuous random variable, which has been shown to be more expressive to capture the uncertainty in real world than discrete time and epoch based models (Xiong et al., 2010; Wang et al., 2015).

**Stochastic Differential Equations (SDEs)**. A SDE is a differential equation in which one or more of the terms is a stochastic process. The SDE models the evolution of state $x_i(t) \in \mathbb{R}$ for user $i$ with a drift, diffusion and jump term:

$$\mathrm{d}x_i(t) = \underset{\text{drift}}{\underset{\uparrow}{f(x_i)\mathrm{d}t}} + \underset{\text{diffusion noise}}{\underset{\uparrow}{g(x_i)\mathrm{d}w_i(t)}} + \underset{\text{jump: point process}}{\underset{\uparrow}{\sum\nolimits_j h(x_j)\mathrm{d}N_j(t)}} \qquad (2)$$

where $\mathrm{d}x_i(t) := x_i(t + \mathrm{d}t) - x_i(t)$ describes the increment of $x_i(t)$. The functions $\{f, g, h\}$ are nonlinear. The drift term captures the evolution of the system; the diffusion term models the noise with the Wiener process, $w_i(t) \sim \mathcal{N}(0, t)$, which follows a Gaussian distribution; the point process $N_j(t)$ models events generated by user $j$ and its intensity $\lambda_j(t)$ is stochastic. The influence function $h(x_j)$ captures social influence, *i.e.*, how user $j$ influences user $i$.

## 3. Intensity Stochastic Control Problem

In this section, we first define the control policy and the controlled stochastic processes; then formulate the stochastic intensity control problem.

**Definition 1** (Controlled Stochastic Processes). *Set $\lambda_i(t)$ as the original (uncontrolled) intensity for $N_i(t)$, $u_i(t) > 0$ as the control policy, and $\tilde{\lambda}_i(u_i(t), t)$ as the controlled intensity of controlled point process $\tilde{N}_i(u_i(t), t)$. The uncontrolled SDE in (2) is modified as the controlled SDE:*

$$\mathrm{d}x_i = f(x_i)\mathrm{d}t + g(x_i)\mathrm{d}w_i + \sum\nolimits_{j=1}^{M} h(x_j)\mathrm{d}\tilde{N}_j(u_j, t) \qquad (3)$$

*For each user i, the form of control policy is:*

$$\tilde{\lambda}_i(u_i(t), t) = \lambda_i(t)u_i(t), \; i = 1, \cdots, M \qquad (4)$$

The control policy $u_i(t)$ helps each user $i$ decide the scale of changes to his original intensity $\lambda_i(t)$ at time $t$, and controls the frequency of generating events. The larger $u_i(t)$, the more likely an event will happen. Moreover, the control policy is in the multiplicative form. The rationale behind this choice is that it makes the policy easy to execute and meaningful in practice. For example, a network moderator may request a user to reduce his tweeting intensity five times if he spreads rumors, or double the original intensity if he posts educational topics. Alternative policy formulations that are based on addition are less intuitive and not easy to execute in practice. For example, if the moderator asks the user to decrease his posting intensity by one, this instruction is difficult to be interpreted in a meaningful way. Finally, since intensity functions are positive, we set $u_i(t) > 0$.

Our goal is to find the best control policy such that this controlled SDE achieves a target state. Next, we formulate the stochastic intensity control problem.

**Definition 2** (Intensity Control Problem)**.** *Given the controlled SDE in (3), the goal is to find* $\boldsymbol{u}^*(t)$ *for* $t \in [0, T]$*, such that the following objective function is minimized:*

$$\boldsymbol{u}^* = \operatorname{argmin}_{\boldsymbol{u}>0} \mathbb{E}_{\boldsymbol{x}} \Big[ S(\boldsymbol{x}) + \gamma C(\boldsymbol{u}) \Big], \qquad (5)$$

*where* $\boldsymbol{x} := \{\boldsymbol{x}(t) | t \in [0, T]\}$ *is the controlled SDE trajectory on* $[0, T]$*,* $\boldsymbol{u}$ *denotes the policy on* $[0, T]$*, The expectation* $\mathbb{E}_{\boldsymbol{x}}$ *is taken over all trajectories of* $\boldsymbol{x}$*, whose stochasticity comes from the Wiener process* $\boldsymbol{w}(t)$ *and controlled point process* $\tilde{\boldsymbol{N}}(\boldsymbol{u}, t)$ *on* $[0, T]$*. The function* $C(\boldsymbol{u})$ *is the control cost, and* $S(\boldsymbol{x})$ *is the state cost defined as follows:*

$$S(\boldsymbol{x}) = \phi(\boldsymbol{x}(T), T) + \int_0^{T^-} q(\boldsymbol{x}(t), t) \mathrm{d}t \qquad (6)$$

*It is a function of the trajectory* $\boldsymbol{x}$ *and measures its cost on* $[0, T]$*.* $q(\boldsymbol{x}(t), t)$ *is the instantaneous state cost at time* $t$*, and* $\phi(\boldsymbol{x}(T), T)$ *is the terminal state cost. The scalar* $\gamma$ *controls the trade-off between state cost and control cost.*

The state cost is a user-defined function and its form depends on different applications. We will provide detailed examples in section 6 later. The control cost captures the budget and effort, such as time and money, to control the system.

## 4. Solution Overview

Directly computing the optimal policy in (5) is difficult using previous control methods (Pham, 1998; Oksendal & Sulem, 2005; Hanson, 2007). The challenges are as follows.

**Challenges**. The first two challenges lie in different problem scopes. First, the control policy in these works is in the drift of SDE, and not directly applicable to the intensity control problem. Second, these works typically consider simple Poisson processes with deterministic intensity. However, in our problem the intensity can also be stochastic, which adds another layer of stochasticity. Besides the problem scopes, these works have two fundamental technical challenges:

**I. Choice of control cost**. These works need to define the form of control cost beforehand, which is nontrivial. For example, $u_i(t) = 1$ means there is no control. However, it is not clear which of the two heuristic forms works better:

$$\int_0^T \|\boldsymbol{u}(t) - \mathbf{1}\|^2, \int_0^T \sum_i (u_i(t) - 1) - \log\big(u_i(t)\big) \mathrm{d}t \quad (7)$$

Unfortunately, prior works need tedious and heuristic tuning of the function forms of control cost $C(\boldsymbol{u})$.

**II. Scalability and approximations**. Prior works rely on the Bellman optimality condition and use stochastic programming to derive the corresponding Hamilton-Jacobi-Bellman (HJB) partial differential equation (PDE). Solving this PDE for multi-dimensional nonlinear SDEs is difficult due to scalability limitations, *i.e.*, *curse of dimensionality* (Hanson, 2007). This is especially challenging in social network applications where the SDE has thousands or millions of dimensions (each user represents one dimension). Efficient solution for the PDE only exists in the special case of *linear* SDE and *quadratic* control cost and state cost. This case is restrictive when the underlying model is a nonlinear SDE, and the state cost is arbitrary function.

**Our approach**. To address the above challenges, we propose a generic framework with the following key steps.

**I. Optimal measure-theoretic formulation**. We establish a novel view of the intensity control problem by linking it to the optimal probability measure. The key insight is to compute the optimal measure $\mathbb{Q}^*$, which is induced by optimal policy $\boldsymbol{u}^*$. With this view, the control cost comes naturally as a KL-divergence term (Section 5.1):

$$\boxed{\mathbb{Q}^* = \operatorname{argmin}_{\mathbb{Q}} \Big[ \mathbb{E}_{\mathbb{Q}}[S(\boldsymbol{x})] + \gamma \mathbb{D}_{KL}(\mathbb{Q}||\mathbb{P}) \Big]}$$

**II. Variational inference for the optimal policy**. It is much easier to find the optimal measure $\mathbb{Q}^*$ compared with directly solving (5). Based on its form, we then parameterize $\mathbb{Q}(\boldsymbol{u})$, and compute $\boldsymbol{u}^*$ by minimizing the distance between $\mathbb{Q}^*$ and $\mathbb{Q}(\boldsymbol{u})$. This approach leads to a scalable and simple algorithm, and does not need any approximations to the nonlinear SDE or cost functions (Section 5.2):

$$\boxed{\boldsymbol{u}^* = \operatorname{argmin}_{\boldsymbol{u}>0} \mathbb{D}_{KL}\big(\mathbb{Q}^* || \mathbb{Q}(\boldsymbol{u})\big)}$$

Finally, we transform the open-loop policy to the feedback policy and develop a scalable algorithm.

## 5. Variational Policy

In this section, we will present technical details of our framework, Variational Policy. We first provide a measure-theoretic view of the control problem, and show that finding optimal measure is equivalent to finding the optimal control. Then we compute the optimal measure and find the optimal control policy from the view of variational inference.

### 5.1. Optimal measure-theoretic formulation of intensity optimal control problem

Each trajectory (sample path) of a SDE is stochastic. Hence we can define a probability measure on all possible trajectories, and a SDE uniquely induces a probability measure. At a conceptual level, the SDE and the measure induced by the SDE are equivalent mathematical representations: obtaining a trajectory from this SDE by simulation (forward propagating the SDE) is equivalent to generating a sample from the probability measure induced by the SDE.

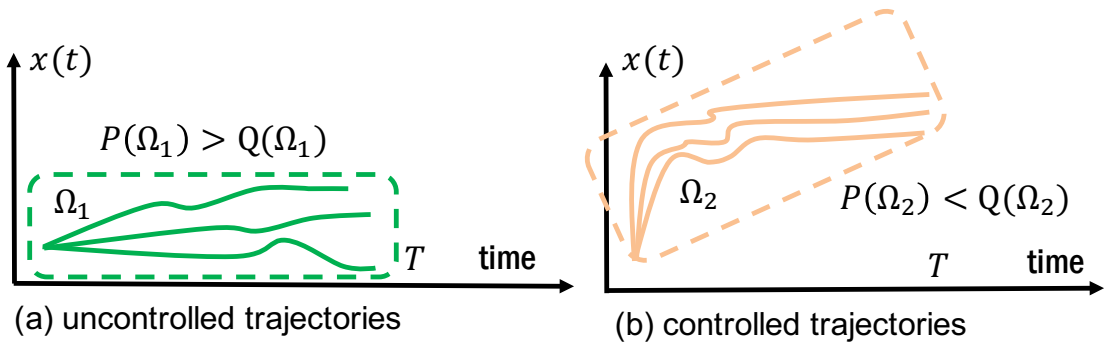

*Figure 2.* Explanation of the measures induced by SDEs. (a) the three green uncontrolled trajectories are in the region of $\Omega_1$. Since $\mathbb{P}$ is induced by the uncontrolled SDE, naturally it has high probability on the region $\Omega_1$ compared with $\mathbb{Q}$. Similarly, the three yellow trajectories are in $\Omega_2$, and $\mathbb{Q}$ has high probability in this region since $\mathbb{Q}$ is induced by the controlled SDE.

Next, we link this probability measure view to the intensity control problem. The problem in (5) aims at finding an optimal policy, which uniquely determines the optimal controlled SDE. Since the SDE induces a measure, (5) is equivalent to the problem of finding the optimal measure.

Mathematically, we set $\mathbb{P}$ as the probability measure induced by the uncontrolled SDE in (2), and set $\mathbb{Q}$ as the measure induced by the controlled SDE in (3). Hence $\mathbb{E}_{\boldsymbol{x}} = \mathbb{E}_{\mathbb{Q}}$, *i.e.*, taking the expectation over stochastic trajectories $\boldsymbol{x}$ in the original objective function is essentially taking expectation over the measure $\mathbb{Q}$. Moreover, the difference between $\mathbb{P}$ and $\mathbb{Q}$ is just the effect of the control policy. Therefore, $\boldsymbol{u}^*$ uniquely induces $\mathbb{Q}^*$. Figure 2 demonstrates $\mathbb{P}$ and $\mathbb{Q}$.

Based on this idea, instead of directly computing $\boldsymbol{u}^*$, we aim at finding the optimal measure $\mathbb{Q}^*$, such that $\mathbb{E}_{\mathbb{Q}}[S(\boldsymbol{x})]$ is minimized. We set the constraint such that $\mathbb{Q}$ is as close to $\mathbb{P}$ as possible, and propose the following objective function:

$$\min_{\mathbb{Q}} \Big[\mathbb{E}_{\mathbb{Q}}[S(\boldsymbol{x})] + \gamma \mathbb{D}_{KL}(\mathbb{Q}||\mathbb{P})\Big], \text{ s.t. } \int \mathrm{d}\mathbb{Q} = 1 \quad (8)$$

where $\int \mathrm{d}\mathbb{Q} = 1$ ensures $\mathbb{Q}$ is a probability measure, and $\mathrm{d}\mathbb{Q}$ is the probability density. $\mathbb{D}_{KL}(\mathbb{Q}||\mathbb{P}) = \mathbb{E}_{\mathbb{Q}}[\log(\frac{\mathrm{d}\mathbb{Q}}{\mathrm{d}\mathbb{P}})]$ is the KL divergence between these two measures.

**Natural control cost.** This KL divergence term provides an elegant way of measuring the distance between controlled and uncontrolled SDEs. Minimizing this term sets $\mathbb{Q}$ to be close to $\mathbb{P}$; hence it provides an implicit measure of the control cost. Mathematically, we express it as follows:

$$\mathbb{D}_{KL}(\mathbb{Q}||\mathbb{P}) := \mathbb{E}_{\mathbb{Q}}[\log(\frac{\mathrm{d}\mathbb{Q}}{\mathrm{d}\mathbb{P}})] \quad (9)$$
$$= \mathbb{E}_{\mathbb{Q}}\Big[\int_0^T \sum_i \Big(\log(u_i(t)) + \frac{1}{u_i(t)} - 1\Big)\lambda_i(t)u_i(t)\mathrm{d}t\Big]$$

Appendix D contains derivations. With this formulation, we set the control cost $C(\boldsymbol{u}) = \log(\frac{\mathrm{d}\mathbb{Q}}{\mathrm{d}\mathbb{P}})$. This function reaches its minimum when $u_i(t) = 1$, since the function $f(x) = (\log(x) + \frac{1}{x} - 1)x$ reaches the minimum when $x = 1$. Interestingly, $C(\boldsymbol{u})$ is none of the heuristics in (7). Hence our control cost comes naturally from the dynamics.

Another benefit of our formulation is that the probability measure that minimizes (8) is easy to derive (Appendix A contains derivations). The optimal measure is

$$\frac{\mathrm{d}\mathbb{Q}^*}{\mathrm{d}\mathbb{P}} = \frac{\exp(-\frac{1}{\gamma}S(\boldsymbol{x}))}{\mathbb{E}_{\mathbb{P}}[\exp(-\frac{1}{\gamma}S(\boldsymbol{x}))]} \quad (10)$$

The term $\frac{\mathrm{d}\mathbb{Q}^*}{\mathrm{d}\mathbb{P}}$ is called the Radon-Nikodym derivative (Dupuis & Ellis, 1997; Theodorou, 2015). This expression is intuitive: if a trajectory $\boldsymbol{x}$ has low state cost, then $\frac{\mathrm{d}\mathbb{Q}^*}{\mathrm{d}\mathbb{P}}$ is large. This means that this trajectory is likely to be sampled from $\mathbb{Q}^*$. In summary, our first contribution is the link between the problem of finding optimal control to that of finding optimal measure. Computing the optimal measure is much easier than directly solving (5).

However, the main challenge in our measure-theoretic formulation is that there is no *explicit* transformation between the optimal measure $\mathbb{Q}^*$ and the optimal control $\boldsymbol{u}^*$. To solve this problem, next we design a convex objective function by matching probability measures.

### 5.2. Finding optimal policy with variational inference

We formulate our objective function based on the optimal measure. More specifically, we find a control $\boldsymbol{u}$ which pushes the induced measure $\mathbb{Q}(\boldsymbol{u})$, as close to the optimal measure as possible. Mathematically, we have:

$$\boldsymbol{u}^* = \operatorname{argmin}_{\boldsymbol{u}>0} \mathbb{D}_{KL}\big(\mathbb{Q}^*||\mathbb{Q}(\boldsymbol{u})\big) \quad (11)$$

From the view of variational inference (Wainwright & Jordan, 2003; Williams et al., 2016), our objective function describes the amount of information loss when $\mathbb{Q}(\boldsymbol{u})$ is used to approximate $\mathbb{Q}^*$. This objective is in sharp contrast to traditional methods that solve the problem by computing the solution of the HJB PDE, which have severe limitations in scalability and feasibility to nonlinear SDEs (Oksendal & Sulem, 2005; Hanson, 2007).

Next, we simplify the objective in (11) and compute the optimal control policy. From the definition of KL divergence and chain rule of derivatives, (11) is expressed as:

$$\mathbb{D}_{KL}(\mathbb{Q}^*||\mathbb{Q}(\boldsymbol{u})) = \mathbb{E}_{\mathbb{Q}^*}\Big[\log\big(\frac{\mathrm{d}\mathbb{Q}^*}{\mathrm{d}\mathbb{P}}\frac{\mathrm{d}\mathbb{P}}{\mathrm{d}\mathbb{Q}(\boldsymbol{u})}\big)\Big]. \quad (12)$$

The derivative $\mathrm{d}\mathbb{Q}^*/\mathrm{d}\mathbb{P}$ is given in (10), and we only need to compute $\mathrm{d}\mathbb{P}/\mathrm{d}\mathbb{Q}(\boldsymbol{u})$. This derivative is the relative density of probability distribution $\mathbb{P}$ w.r.t. $\mathbb{Q}(\boldsymbol{u})$. The change of probability measure happens because the intensity is changed from $\boldsymbol{\lambda}(t)$ to $\tilde{\boldsymbol{\lambda}}(\boldsymbol{u}, t)$. Hence $\mathrm{d}\mathbb{P}/\mathrm{d}\mathbb{Q}(\boldsymbol{u})$ is essentially the *likelihood ratio* between the uncontrolled and controlled point process. We summarize its form in Theorem 3.

**Theorem 3.** *For the intensity control problem, we have:* $\mathrm{d}\mathbb{P}/\mathrm{d}\mathbb{Q}(\boldsymbol{u}) = \exp\big(\mathcal{D}(\boldsymbol{u})\big)$*, where* $\mathcal{D}(\boldsymbol{u})$ *is expressed as:*

$$\sum\nolimits_{i=1}^{M} \int_0^T \big(u_i(s) - 1\big)\lambda_i(s)\mathrm{d}s - \int_0^T \log\big(u_i(s)\big)\mathrm{d}N_i(s)$$

Appendix B contains details of the proof. Next we substitute $\mathrm{d}\mathbb{Q}^*/\mathrm{d}\mathbb{P}$ and $\mathrm{d}\mathbb{P}/\mathrm{d}\mathbb{Q}(\boldsymbol{u})$ to (12). After removing terms independent of $\boldsymbol{u}$, the objective function is simplified as:

$$\boldsymbol{u}^* = \mathrm{argmin}_{\boldsymbol{u}>0}\, \mathbb{E}_{\mathbb{Q}^*}[\mathcal{D}(\boldsymbol{u})]$$

Next, we will solve this optimization problem to compute $\boldsymbol{u}^*$. As in traditional stochastic optimal control works (Oksendal & Sulem, 2005; Hanson, 2007), a control policy is obtained by solving the HJB PDE at discrete timestamps on $[0, T]$. Hence it suffices to parameterize our policy $\boldsymbol{u}(t)$ as a piecewise constant function on $[0, T]$.

We denote the $k$-th piece of $\boldsymbol{u}$ as $\boldsymbol{u}^k$, which is defined on $[k\Delta t, (k+1)\Delta t)$, with $k = 0, \cdots, K-1$, $t_k = k\Delta t$ and $T = t_K$. Now we express the objective function as follows.

$$\mathbb{E}_{\mathbb{Q}^*}[\mathcal{D}(\boldsymbol{u})] = \sum\nolimits_i \sum\nolimits_k \Big(\mathbb{E}_{\mathbb{Q}^*}\big[\int_{t_k}^{t_{k+1}} (u_i^k - 1)\lambda_i(s)\mathrm{d}s\big] - \mathbb{E}_{\mathbb{Q}^*}\big[\int_{t_k}^{t_{k+1}} \log(u_i^k)\mathrm{d}N_i(s)\big]\Big) \quad (13)$$

where $u_i^k$ denotes the $i$-th dimension of $\boldsymbol{u}^k$. We just need to focus on the parts that involves $u_i^k$ and move it outside of the expectation. Further we can show the final expression is convex in $u_i^k$. Finally, setting the gradient to zero yields the following optimal control policy, denoted as $u_i^{k*}$:

$$u_i^{k*} = \frac{\mathbb{E}_{\mathbb{P}}\big[\exp(-\frac{1}{\gamma}S(\boldsymbol{x}))\int_{t_k}^{t_{k+1}}\mathrm{d}N_i(s)\big]}{\mathbb{E}_{\mathbb{P}}\big[\exp(-\frac{1}{\gamma}S(\boldsymbol{x}))\int_{t_k}^{t_{k+1}}\lambda_i(s)\mathrm{d}s\big]} \quad (14)$$

Appendix C contains complete derivations. Note we have transformed $\mathbb{E}_{\mathbb{Q}^*}$ to $\mathbb{E}_{\mathbb{P}}$ using (10). It is important because $\mathbb{E}_{\mathbb{Q}^*}$ is not directly computable. Inspired by the idea of importance sampling, since we only know the SDE of the uncontrolled dynamics in (2) and can only compute the expectation under $\mathbb{P}$, the change of expectation is necessary.

To compute $\mathbb{E}_{\mathbb{P}}$, we use the Monte Carlo method to sample $I$ trajectories from (2) on $[0, T]$ and take the sample average. To obtain the $m$-th sample $\boldsymbol{x}^m$, we use the classic routine: sample point process $\boldsymbol{N}^m(t)$ (*e.g.*, Hawkes process) using thinning algorithm (Ogata, 1981), sample Wiener process $\boldsymbol{w}^m(t)$ from Gaussion distribution, and apply the Euler method (Hanson, 2007) to obtain $\boldsymbol{x}^m$. Since each sample is independent, it can be scaled up easily with parallelization.

Next, we compute $w^m = \exp(-S(\boldsymbol{x}^m)/\gamma)$ by evaluating the state cost, and compute $\int_{t_k}^{t_{k+1}} \mathrm{d}N_i^m(s)$ as the number

**Algorithm 1** KL - Model Predictive Control

**Input:** sample size $I$, optimization window length $\tilde{T}$, total time window $T$, timestamps $\{t_k\}$ on $[0, T]$.
**Output:** optimal control $\boldsymbol{u}^*$ at each $t_k$ on $[0, T]$.
**for** $k = 0$ **to** $K - 1$ **do**
  **for** $m = 1$ **to** $I$ **do**
    Sample $\mathrm{d}\boldsymbol{N}(t)$, $\mathrm{d}\boldsymbol{w}(t)$ and generate $\boldsymbol{x}^m$ on $[t_k, t_k + \tilde{T}]$ according to (2) and the current state.
    $S(\boldsymbol{x}^m) = \int_0^T q(\boldsymbol{x}^m)\mathrm{d}t + \phi(\boldsymbol{x}^m)$, $w^m = \exp(-\frac{1}{\gamma}S)$
  **end for**
  Compute $u_i^{k*}$ from (15) for each $i$, and execute $\boldsymbol{u}^{k*}$, receive state feedback and update state.
**end for**

of events that occurred during $[t_k, t_{k+1})$ at the $i$-th dimension. Moreover, since $\lambda_i^m(t)$ is history-dependent, given the events history in the $m$-th sample, $\lambda_i^m(t)$ is fixed with a parametric form. Hence $\int_{t_k}^{t_{k+1}} \lambda_i^m(s)\mathrm{d}s$ can also be computed numerically or in closed form. The closed form expression exists for the Hawkes process. In summary, the sample average approximation of (14) is:

$$u_i^{k*} = \frac{\sum_{m=1}^{I} w^m \int_{t_k}^{t_{k+1}} \mathrm{d}N_i^m(s)}{\sum_{m=1}^{I} w^m \int_{t_k}^{t_{k+1}} \lambda_i^m(s)\mathrm{d}s} \quad (15)$$

Next, we discuss the properties of our policy.

**Stochastic intensity**. The intensity function $\lambda_i(t)$ is history independent and stochastic, *e.g.*, Hawkes process. Since $\lambda_i(t)$ is inside the expectation $\mathbb{E}_{\mathbb{P}}$ in (14), our policy naturally considers its stochasticity by taking the expectation.

**General SDE & arbitrary cost**. Since we only need the SDE system to sample trajectories, our framework is applicable to general nonlinear SDEs and arbitrary cost functions.

### 5.3. From open-loop policy to feedback policy

The current control policy in (15) does not depend on the system's feedback. However, a more effective policy should consider the current state of SDE, and integrate such feedback into the policy. In this section, we will transform the open-loop policy into a feedback policy.

To design this feedback policy, we use the model predictive control (MPC) scheme (Camacho & Alba, 2013), where the *Model* of the process is used to *Predict* the future evolution of the process to optimize the *Control*. In MPC, online optimization and execution are interleaved as follows.

*(i)* *Optimization.* At time $t$, we compute the control policy $\boldsymbol{u}^*$ on $[t, t+\tilde{T}]$ using (15) for a short time horizon $\tilde{T} \ll T$ in the future. Therefore, we only need to sample trajectories on $[t, t+\tilde{T}]$ for computation instead of $[0, T]$.

*(ii)* *Execution.* We apply the first optimal move $\boldsymbol{u}^*(t)$ at this time $t$, and observe the new system state.

*(iii)* *Feedback & re-optimization.* At time $t+1$, with the new observed state, we re-compute the control and repeat the above process. Algorithm 1 summarizes the procedure.

The advantage of MPC is that it yields a *feedback* control that implicitly depends on the current state $\boldsymbol{x}(t)$. Moreover, separating the optimization horizon $\tilde{T}$ from $T$ is also advantageous since it makes little sense to consider choosing a deterministic set of actions far out into the future.

# 6. Applications

In this section, we apply our framework to two real-world applications in social sciences.

**Guiding opinion diffusion**. The continuous-time opinion model considers the opinion and timing of each posting event (De et al., 2015; He et al., 2015). It assigns each user $i$ a Hawkes intensity $\lambda_i(t)$ and an opinion process $x_i(t) \in \mathbb{R}$ where $x_i(t) = 0$ corresponds to neutral opinion. Users are connected according to a network adjacency matrix $\boldsymbol{A} = (\alpha_{ij})$. The opinion change of user is captured by three terms:

$$\mathrm{d}x_i(t) = \big(b_i - x_i\big)\mathrm{d}t + \beta \mathrm{d}w_i(t) + \sum\nolimits_j \alpha_{ij} x_j \mathrm{d}N_j(t) \quad (16)$$

where $b_i$ is the baseline opinion, *i.e.*, personal characteristics. The noise process $\mathrm{d}w_i(t)$ captures the normal fluctuations in the dynamics due to unobserved factors such as activity outside the social platform and unexpected events. The jump term captures the fact that the change of user $i$'s opinion is a weighted summation of his neighbors' influence, and $\alpha_{ij}$ ensures only the opinion of a user's neighbor is considered.

How to control users' posting intensity, such that the opinion dynamics is steered towards a target? We can modify each user's opinion posting process $N_j(t)$ as $\tilde{N}_j(u_j, t)$ with policy $u_j(t)$. Common choices of state costs are as follows:

- *Least square opinion shaping.* The goal is to make the expected opinion to achieve the target $\boldsymbol{a}$, *e.g.*, nobody believes the rumor during the period. Mathematically, we set $q = \|\boldsymbol{x}(t) - \boldsymbol{a}\|^2$ and $\phi = \|\boldsymbol{x}(T) - \boldsymbol{a}\|^2$.
- *Opinion influence maximization.* The goal is to maximize each user's positive opinion, *e.g.*, a political party maximizes the support during the election period. Mathematically, we set $q = -\sum_i x_i(t)$ and $\phi = -\sum_i x_i(T)$.

**Guiding broadcasting behavior**. When a user posts in social network, he competes with others that his followers follow, and he will gain greater attention if his posts remain top among followers' feeds. His position defined as the rank of his post among his followers. (Karimi et al., 2016) models the change of a broadcaster's position due to the posting behavior of other competitors and himself as follows.

$$\mathrm{d}x_j(t) = \mathrm{d}N_o(t) - \big(x_j(t) - 1\big)\mathrm{d}N_i(t) \quad (17)$$

where $i$ is the broadcaster and $j \in \mathcal{F}(i)$ denote one follower of $i$. The stochastic process $x_j(t) \in \mathbb{N}$ denotes the rank of broadcaster $i$'s posts among all the posts that his follower $j$ receives. Rank $x_j = 1$ means $i$'s posts is the top-1 among all posts $j$ receives. $N_i(t)$ is a Poisson process capturing the broadcaster's posting behavior. $N_o(t)$ is the Hawkes process for the behavior of all other broadcasters that $j$ follows.

How to change the posting intensity of the broadcaster, such that his posts always remain on top? We use the policy to change $N_i(t)$ to $\tilde{N}_i(u_i, t)$ and help user $i$ decide when to post messages. The state cost minimizes his rank among all followers' news feed. Specifically, we set the state and terminal cost as $q = \sum_{j \in \mathcal{F}(i)} x_j(t)$ and $\phi = \sum_{j \in \mathcal{F}(i)} x_j(T)$.

# 7. Experiments

We focus on two applications in the previous section: least square opinion guiding and smart broadcasting. We compare with suitable stochastic optimization approaches that are popular in reinforcement learning and heuristics.

- Cross Entropy (CE) (Stulp & Sigaud, 2012): It samples controls from a Gaussian distribution, sorts the samples in ascending order with respect to the cost and recomputes the distribution parameters based on the first $K$ elite samples. Then it returns to the first step with a new distribution until the cost converges.
- Finite Difference (FD) (Peters & Schaal, 2006): It generates $I$ samples of perturbed policies $\boldsymbol{u} + \Delta\boldsymbol{u}$ and computes perturbed cost $S + \Delta S$. Then it uses them to approximate the true gradient of the cost with respect to the policy.
- Greedy: It controls the system when local state cost is high. We divide the window into $n$ state cost observation timestamps. At each timestamp, Greedy computes state cost and controls the system based on pre-specified control rules if current cost is more than $k$ times of the optimal cost of our algorithm. It will stop if it has reached the current budget bound. We vary $k$ from 1 to 5, $n$ from 1 to 100 and report the best performance.
- Base Intensity (BI) (Farajtabar et al., 2014): It sets the policy for the base parameterization of the intensity only at initial time and does not consider the system feedback.

We provide both MPC and open-loop (OL) versions for our KL algorithm, Finite Difference and Cross Entropy. For MPC, we set the optimization window $\tilde{T} = T/10$ and sample size $I = 10,000$. It is efficient to generate these samples and takes less than one second using parallelization.

## 7.1. Experiments on Opinion Guiding

We generate a synthetic network with 1000 users. We simulate the opinion SDE on window $[0, 50]$ by applying Euler forward method (Süli & Mayers, 2003) to compute the difference form of the SDE in (2). The time window is divided into 500 timestamps. We set the initial opinion $x_i(0) = -10$ and the target opinion $a_i = 1$ for each user. For model parameters, we set $\beta = 0.2$, and adjacency matrix $\boldsymbol{A}$ generated uniformly on $[0, 0.01]$ with sparsity of $0.001$. We simulate the Hawkes process using thethinning algorithm (Ogata,

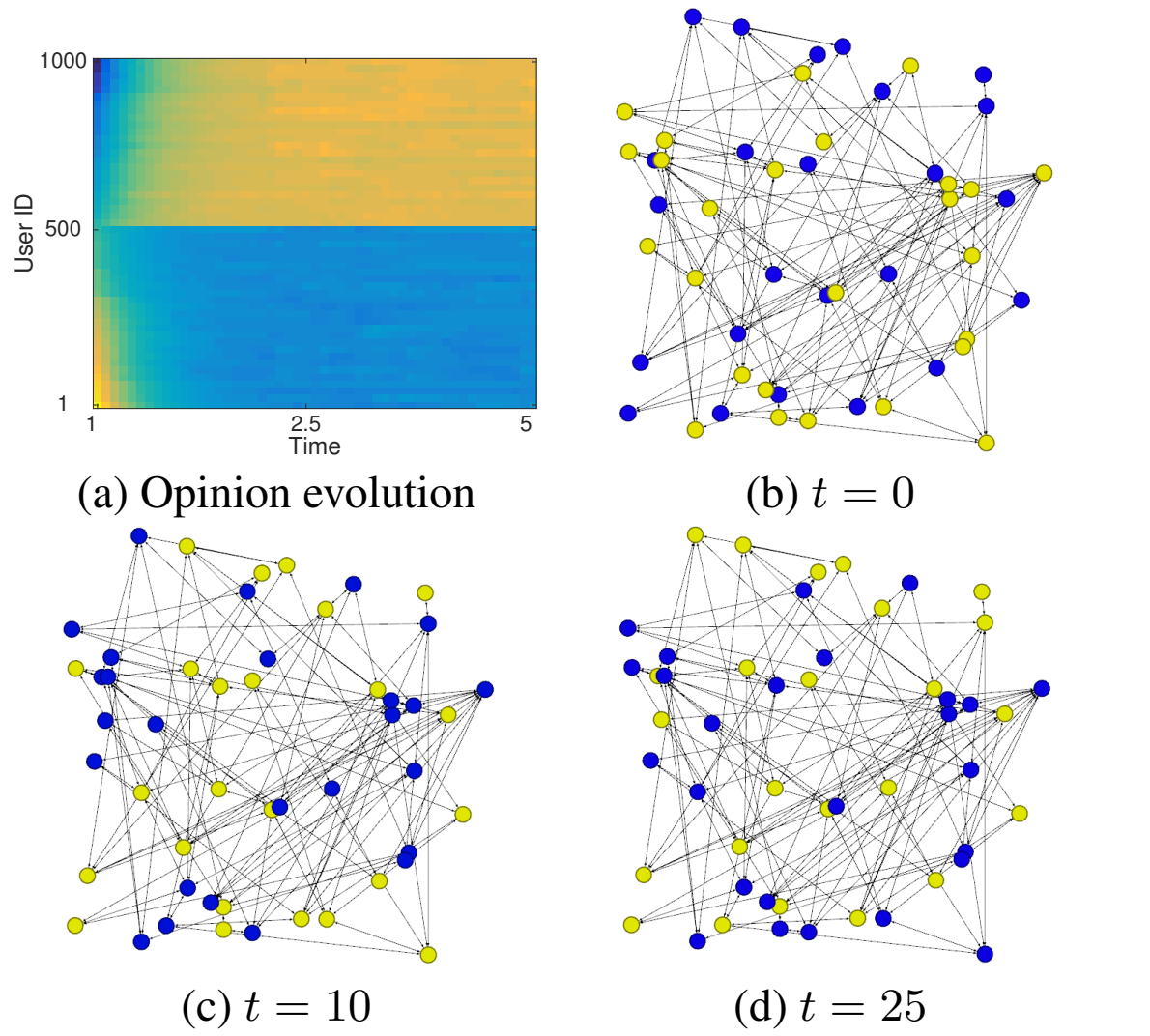

(a) Opinion evolution (b) $t = 0$ (c) $t = 10$ (d) $t = 25$

*Figure 3.* Controlled opinion dynamics of 1000 users. The initial opinions are uniformly sampled from $[-10, 10]$ and sorted, target opinion $\boldsymbol{a}$ is polarized with $-5$ and 10. (a) shows the opinion *value* per user over time. (b-d) are network snapshots of the opinion *polarity* of 50 sub-users. Yellow/blue means positive/negative.

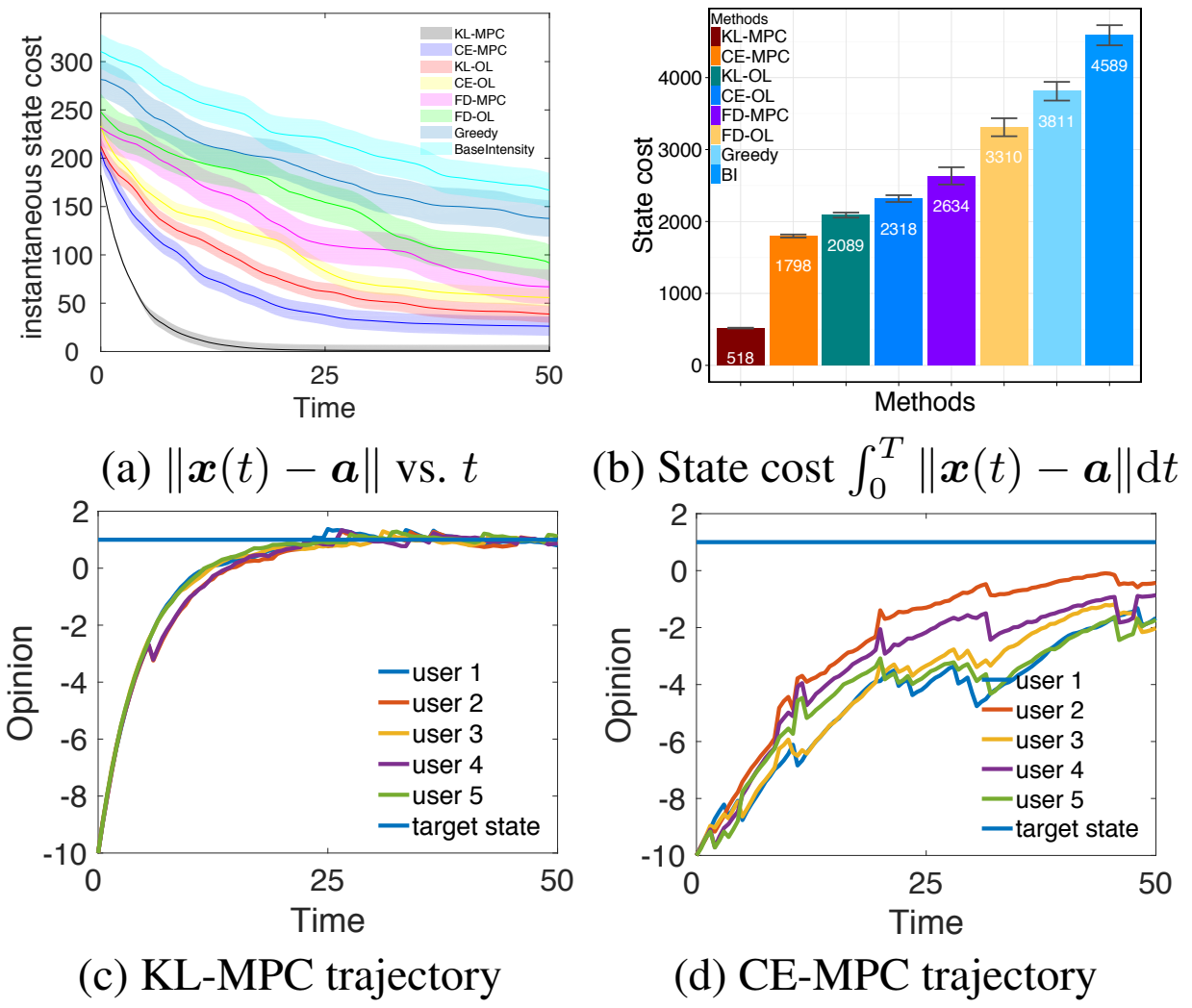

(a) $\|\boldsymbol{x}(t) - \boldsymbol{a}\|$ vs. $t$ (b) State cost $\int_0^T \|\boldsymbol{x}(t) - \boldsymbol{a}\| \mathrm{d}t$ (c) KL-MPC trajectory (d) CE-MPC trajectory

*Figure 4.* Experiments on least square guiding. (a) Instantaneous cost vs. time. Line is the mean and pale region is the variance; (b) state cost; (c,d) sample opinion trajectories of five users.

1981). We set the base intensity in (1) to be $\boldsymbol{\mu} = \mathbf{0.01}$; the influence matrix is the same as the adjacency matrix $\boldsymbol{A}$. We set the cost tradeoff parameter to be $\gamma = 10$.

Figure 3 shows the controlled opinion at different times. Our method works efficiently with fast convergence speed. Figure 4(a) shows the instantaneous cost $\|\boldsymbol{x}(t) - \boldsymbol{a}\|$ at each time $t$. The opinion system is gradually steered towards the target, and the cost decreases over time. Our KL-MPC achieves the lowest instantaneous cost at each time and has the fastest convergence to the optimal cost. Hence the total state cost is also the lowest.

Figure 4(b) shows that KL-MPC has $3\times$ cost improvement than CE-MPC, with less variance and faster convergence. This is because KL-MPC is more flexible and has less restrictions on the control policy. CE-MPC is a popular method for the traditional control problem in robotics, where the SDE does not contain the jump term and control is in the drift. However, CE-MPC assumes the control is sampled from a Gaussian distribution, which might not be the ideal assumption in the intensity control problem. FD performs worse than CE due to the error in the gradient estimation process. Finally, for the same method, the MPC always performs better than open-loop version, which shows the importance of incorporating state feedback to the policy.

Figure 5(a,b) compare the controlled intensity with the uncontrolled intensity at the beginning. Since the goal is to influence everyone to be positive, (a) shows that if the user tweets positive opinion, the control will increase its intensity to influence others positively. On the contrary, (b) shows that if the user's opinion is negative, his intensity will be controlled to be small. (c) and (d) show scenarios near the terminal time. Since the system is around the target state, the policy is small and the original and controlled intensity are similar for positive and negative users.

### 7.2. Experiments on Smart Broadcasting

We evaluate on a real-world Twitter dataset (Farajtabar et al., 2015), which contains 280,000 users with 550,000 tweets/retweets. We first learn the parameters of the point processes that capture each user's posting behavior by maximizing the likelihood function of data (Karimi et al., 2016). For each broadcaster, we track down all followers and record all the tweets they posted and reconstruct followers' timelines by collecting all the tweets by people they follow.

We use two evaluation schemes. First, similar to the synthetic case, with learned parameters, we simulate posting events on $[0, 10]$ and conduct control over the simulated dynamics with the cost tradeoff parameter as $\gamma = 10$. The time window is divided into ten timestamps. We repeat this simulation procedure ten times.

The second and more interesting scheme is to carry the policy in a real platform. Since it is very challenging to do so, we mimic it using held-out data. We partition the data into ten intervals and use one interval for training and others for testing. Each method essentially predicts which interval has smaller cost, by measuring the optimal position computed from that method to real position. Specifically, for each broadcaster, the procedure is as follows: (i) Estimate model parameters using data in interval 1. (ii) Compute the optimal policy and obtain the broadcaster's optimal position $x_i^*$ in each other interval $i$. Then sort intervals according to $|x_i - x_i^*|$. (iii) Sort intervals according to the actual value

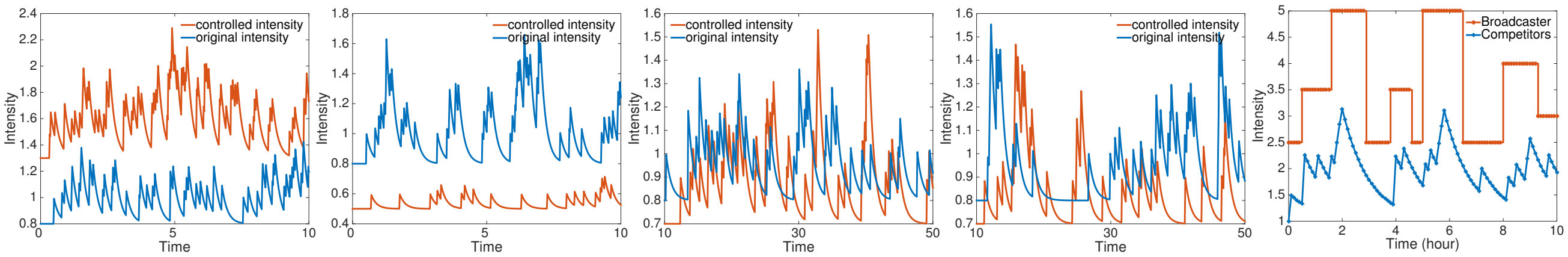

(a) POS user on $[0, 10]$ (b) NEG user on $[0, 10]$ (c) POS user on $[10, 50]$ (d) NEG user on $[10, 50]$ (e) broadcaster vs. others

*Figure 5.* Intensity comparison. (a-d) Opinion guiding experiment: visualization for users with positive (POS) and negative (NEG) opinions during different periods. (e) Smart broadcasting: visualization for one randomly picked broadcaster and his competitors.

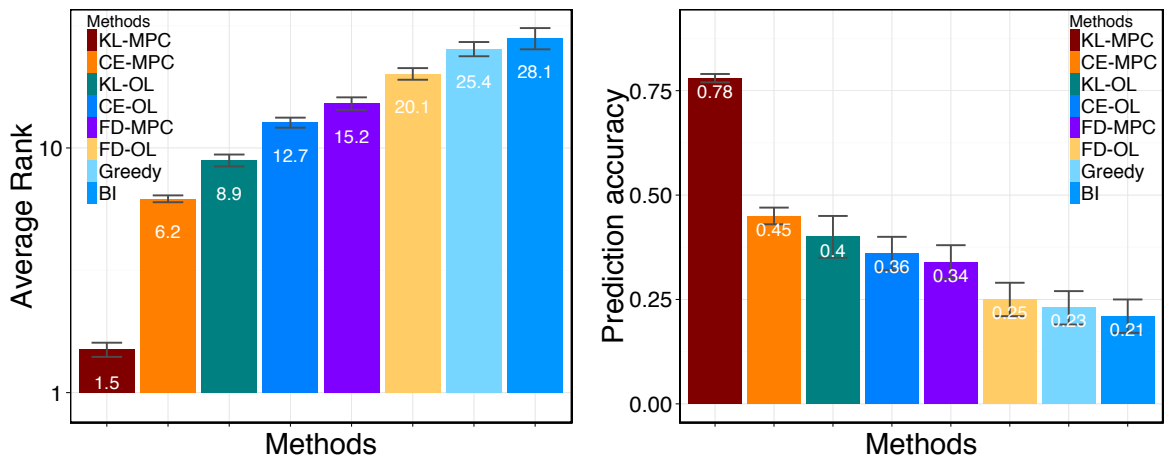

(a) State cost (average rank) (b) prediction accuracy

*Figure 6.* Real world experiment with two evaluation schemes.

of $x_i$. (iv) Compute *prediction accuracy* by dividing the number of pairs with consistent ordering in step 2 and step 3 by total number of pairs. We report the accuracy over ten runs by choosing each different interval for training once.

Figure 6(a) compares the average rank of the broadcaster of different methods. We compute the average rank by dividing the state cost by window length, and average over all broadcasters. KL-MPC achieves the lowest average rank and is $4\times$ lower than the CE-MPC. Specifically, it achieves the rank around $1.5$ at each time, which is nearly the ideal scenario where the broadcaster always remains on top-1. Figure 6(b) further shows that our method performs the best: it achieves more than $0.3+$ improvement over CE-MPC; hence our method has 30% more of the total realizations correctly. Accurate prediction means that if applying our control policy to the users, we will achieve the objective much better than alternative methods.

Figure 5(e) compares the controlled intensity of one broadcaster with the uncontrolled intensity of his competitors. It shows that KL-MPC increases his intensity when that of other competitors is large, and decreases his intensity when competitor's intensity is small. For example, around timestamp 2 and 4, competitors have large intensities; hence to remain on top, this broadcaster needs to double his intensity to create more posts. Moreover, on $[6.5, 8]$, others are not active and this broadcaster keeps a low intensity. His behavior is adaptive since our control cost ensures the broadcaster not to deviate too much from his original intensity.

## 8. Further Related Work

We first review relevant works in the machine learning community. Some works focus on controlling the point process itself, but they are not generalizable for two reasons: (i) the processes are simple, such as Poisson process (Brémaud, 1981) and a power-law decaying function (Bayraktar & Ludkovski, 2014); (ii) the systems only contain point process. However, in social sciences, the system can be driven by many other stochastic processes. Based on Hawkes process, (Farajtabar et al., 2014) designed its baseline intensity to achieve a steady state behavior. However, this policy does not incorporate system feedback. Recently, (Zarezade et al., 2017) proposed to control a user's posting intensity, which is driven by a homogeneous Poisson process. The intensity of this user's competitors is driven by Hawkes processes, and the SDE system has linear coefficients. This method computes the optimal policy by solving a HJB PDE.

In the area of stochastic optimal control, a relevant line of research focuses on event triggered control (Ades et al., 2000; Lemmon, 2010; Heemels et al., 2012; Meng et al., 2013). But the problem is different: their system is linear and only contains a diffusion process, with the control affine in drift and updated at event time. The event times are driven by a fixed point process. However, we study jump diffusion SDEs and directly control the intensity that drives the time of event. Hence our work is unique among previous works.

## 9. Conclusions

We have presented a generic framework to control the stochastic intensity function of a general point process, such that a nonlinear SDE driven by the point process is steered towards a target state. We exploit the measure-theoretic view of the stochastic intensity control problem, derive an analytical form of the optimal measure, and compute the optimal policy using a KL divergence objective. We provide a scalable algorithm with superior performance in diverse social problems. There are many interesting venues for future work. For example, we can apply our method to other interesting problems, such as influence and activity maximization (Kempe et al., 2003; Farajtabar et al., 2014).

**Acknowledgements**. This project was supported in part by NSF IIS-1218749,NIH BIGDATA 1R01GM108341, NSF CAREER IIS-1350983, NSF IIS-1639792 EAGER, ONR N00014-15-1-2340, NVIDIA, Intel and Amazon AWS.

## A. Derivations of the Optimal Measure

The problem of finding the optimal measure is as follows:

$$\min_{\mathbb{Q}} \Big[\mathbb{E}_{\mathbb{Q}}[S(\boldsymbol{x})] + \gamma \mathbb{D}_{KL}(\mathbb{Q}||\mathbb{P})\Big], \text{ s.t. } \int \mathrm{d}\mathbb{Q} = 1 \tag{18}$$

The minimum in (18) is attained at optimal measure $\mathbb{Q}^*$ given by:

$$\frac{\mathrm{d}\mathbb{Q}^*}{\mathrm{d}\mathbb{P}} = \frac{\exp(-\frac{1}{\gamma}S(\boldsymbol{x}))}{\mathbb{E}_{\mathbb{P}}[\exp(-\frac{1}{\gamma}S(\boldsymbol{x}))]} \tag{19}$$

Next, we show the derivations of (19), which contain two parts. First, we will show the following inequality:

$$\gamma \log\left(\mathbb{E}_{\mathbb{P}}\Big[\exp\big(-\frac{1}{\gamma}S(\boldsymbol{x})\big)\Big]\right) \leqslant \Big[\mathbb{E}_{\mathbb{Q}}[S(\boldsymbol{x})] + \gamma \mathbb{D}_{KL}(\mathbb{Q}||\mathbb{P})\Big] \tag{20}$$

The second part is to show the minimum of the above inequality is reached at (19).

To prove the first part, we first express $\mathbb{E}_{\mathbb{P}}$ in the left-hand-side of (20) as a function of the expectation $\mathbb{E}_{\mathbb{Q}}$. More specifically, we have:

$$\log\left(\mathbb{E}_{\mathbb{P}}\Big[\exp\big(-\frac{1}{\gamma}S(\boldsymbol{x})\big)\Big]\right) = \log\left(\int \exp\big(-\frac{1}{\gamma}S(\boldsymbol{x})\big)\mathrm{d}\mathbb{P}\right) \tag{21}$$

$$= \log\left(\int \exp\big(-\frac{1}{\gamma}S(\boldsymbol{x})\big)\frac{\mathrm{d}\mathbb{P}}{\mathrm{d}\mathbb{Q}}\mathrm{d}\mathbb{Q}\right) \tag{22}$$

$$\geqslant \int \log\left(\exp\big(-\frac{1}{\gamma}S(\boldsymbol{x})\big)\frac{\mathrm{d}\mathbb{P}}{\mathrm{d}\mathbb{Q}}\right)\mathrm{d}\mathbb{Q} \tag{23}$$

where (23) is due to the Jensen's inequality that puts the $\log$ operator inside the integral. The measure $\mathbb{P}$ is absolute continuous with respect to $\mathbb{Q}$, hence the derivative $\frac{\mathrm{d}\mathbb{P}}{\mathrm{d}\mathbb{Q}}$ exists.

Moreover, using the property that $\log(ab) = \log a + \log b$ and $\log(1/a) = -\log a$, the right-hand-side of the above inequality can be written as:

$$\begin{aligned}\int \log\left(\exp\big(-\frac{1}{\gamma}S(\boldsymbol{x})\big)\frac{\mathrm{d}\mathbb{P}}{\mathrm{d}\mathbb{Q}}\right)\mathrm{d}\mathbb{Q} &= \int \left(-\frac{1}{\gamma}S(\boldsymbol{x}) + \log\frac{\mathrm{d}\mathbb{P}}{\mathrm{d}\mathbb{Q}}\right)\mathrm{d}\mathbb{Q} \\ &= \int -\frac{1}{\gamma}S(\boldsymbol{x})\mathrm{d}\mathbb{Q} + \int \log\frac{\mathrm{d}\mathbb{P}}{\mathrm{d}\mathbb{Q}}\mathrm{d}\mathbb{Q} \\ &= \int -\frac{1}{\gamma}S(\boldsymbol{x})\mathrm{d}\mathbb{Q} - \int \log\frac{\mathrm{d}\mathbb{Q}}{\mathrm{d}\mathbb{P}}\mathrm{d}\mathbb{Q} \\ &= -\frac{1}{\gamma}\mathbb{E}_{\mathbb{Q}}[S(\boldsymbol{x})] - \mathbb{D}_{\mathbb{KL}}(\mathbb{Q}||\mathbb{P})\end{aligned} \tag{24}$$

Hence, combining (23) and (24), we have:

$$\log\left(\mathbb{E}_{\mathbb{P}}\Big[\exp\big(-\frac{1}{\gamma}S(\boldsymbol{x})\big)\Big]\right) \geqslant -\frac{1}{\gamma}\mathbb{E}_{\mathbb{Q}}[S(\boldsymbol{x})] - \mathbb{D}_{\mathbb{KL}}(\mathbb{Q}||\mathbb{P}) \tag{25}$$

Finally, since $\gamma > 0$, multiply both sides of (25) by $-\gamma$ yields:

$$-\gamma \log\left(\mathbb{E}_{\mathbb{P}}\Big[\exp\big(-\frac{1}{\gamma}S(\boldsymbol{x})\big)\Big]\right) \leqslant \mathbb{E}_{\mathbb{Q}}[S(\boldsymbol{x})] + \gamma \mathbb{D}_{\mathbb{KL}}(\mathbb{Q}||\mathbb{P}) \tag{26}$$

This finishes the proof of (20), the first part of the theorem. Next, we will show the minimum is reached at $\mathbb{Q}^*$ given by (19).

To prove the second part, we will substitute (19) to the right-hand-side of (25) to show that the infimum is reached with this $\mathbb{Q}^*$. More specifically,

$$
\begin{aligned}
\mathbb{E}_{\mathbb{Q}^*}[S(\boldsymbol{x})] + \gamma \mathbb{D}_{\mathbb{KL}}(\mathbb{Q}^*||\mathbb{P}) &= \mathbb{E}_{\mathbb{Q}^*}[S(\boldsymbol{x})] + \gamma \int \log \frac{\mathrm{d}\mathbb{Q}^*}{\mathrm{d}\mathbb{P}} \mathrm{d}\mathbb{Q}^* \\
&= \mathbb{E}_{\mathbb{Q}^*}[S(\boldsymbol{x})] + \gamma \int \log \frac{\exp(-\frac{1}{\gamma}S(\boldsymbol{x}))}{\mathbb{E}_{\mathbb{P}}[\exp(-\frac{1}{\gamma}S(\boldsymbol{x}))]} \mathrm{d}\mathbb{Q}^* \\
&= \mathbb{E}_{\mathbb{Q}^*}[S(\boldsymbol{x})] + \gamma \int -\frac{1}{\gamma} S(\boldsymbol{x}) \mathrm{d}\mathbb{Q}^* - \gamma \int \log \Big( \mathbb{E}_{\mathbb{P}} \Big[ \exp\big(-\frac{1}{\gamma}S(\boldsymbol{x})\big) \Big] \Big) \mathrm{d}\mathbb{Q}^* \qquad (27) \\
&= \mathbb{E}_{\mathbb{Q}^*}[S(\boldsymbol{x})] - \int S(\boldsymbol{x}) \mathrm{d}\mathbb{Q}^* - \gamma \log \Big( \mathbb{E}_{\mathbb{P}} \Big[ \exp\big(-\frac{1}{\gamma}S(\boldsymbol{x})\big) \Big] \Big) \int \mathrm{d}\mathbb{Q}^* \\
&= \mathbb{E}_{\mathbb{Q}^*}[S(\boldsymbol{x})] - \mathbb{E}_{\mathbb{Q}^*}[S(\boldsymbol{x})] - \gamma \log \Big( \mathbb{E}_{\mathbb{P}} \Big[ \exp\big(-\frac{1}{\gamma}S(\boldsymbol{x})\big) \Big] \Big) \qquad (28) \\
&= -\gamma \log \Big( \mathbb{E}_{\mathbb{P}} \Big[ \exp\big(-\frac{1}{\gamma}S(\boldsymbol{x})\big) \Big] \Big)
\end{aligned}
$$

where (27) is due to the property $\log(a/b) = \log a - \log b$ and (28) is because $\mathbb{Q}^*$ is a probability measure hence $\int \mathrm{d}\mathbb{Q}^* = 1$. Hence the infimum is reached and this finishes the proof of the second part.

## B. Proof of Theorem 3

**Theorem 3.** *For the intensity control problem in (4), we have:* $\frac{\mathrm{d}\mathbb{P}}{\mathrm{d}\mathbb{Q}(\boldsymbol{u})} = \exp\big(\mathcal{D}(\boldsymbol{u})\big)$, *where* $\mathcal{D}(\boldsymbol{u})$ *is expressed as*

$$\sum_{i=1}^{M} \int_0^T \big(u_i(s) - 1\big)\lambda_i(s)\mathrm{d}s - \int_0^T \log\big(u_i(s)\big)\mathrm{d}N_i(s)$$

*Proof.* Intuitively, the derivative $\mathrm{d}\mathbb{P}/\mathrm{d}\mathbb{Q}(\boldsymbol{u})$ means the relative density of probability distribution $\mathbb{P}$ with respect to $\mathbb{Q}$. The change of probability measure happens because the intensity of the point process that drives the SDE in (2) is changed from $\boldsymbol{\lambda}(t)$ to $\boldsymbol{\lambda}(\boldsymbol{u}, t)$ in (4). Hence $\mathrm{d}\mathbb{P}/\mathrm{d}\mathbb{Q}(\boldsymbol{u})$ describes the change of probability measure for point processes and is the *likelihood ratio* between the uncontrolled and controlled point process (Brémaud, 1981):

$$\frac{\mathrm{d}\mathbb{P}}{\mathrm{d}\mathbb{Q}(\boldsymbol{u})} = \frac{\exp\big(\mathcal{L}(\boldsymbol{\lambda})\big)}{\exp\big(\mathcal{L}(\boldsymbol{\lambda}(\boldsymbol{u}))\big)} = \exp\big(\mathcal{D}(\boldsymbol{u})\big),$$

where $\mathcal{L}$ is the *log-likelihood* for the multi-dimension point process with $\mathcal{L}(\boldsymbol{\lambda}) = \sum_{i=1}^{M} \mathcal{L}(\lambda_i)$. It is defined as the summation of log-likelihood $\mathcal{L}(\lambda_i)$ of each dimension $i$, where $\mathcal{L}(\lambda_i)$ is defined as follows (Aalen et al., 2008):

$$\mathcal{L}(\lambda_i(t)) = \int_0^T \log(\lambda_i(t))\mathrm{d}N_i(t) - \int_0^T \lambda_i(t)\mathrm{d}t \tag{29}$$

where the operation $\int f(t)\mathrm{d}N(t)$ is defined as the summation of the value of function $f$ at each event time: $\int f(t)\mathrm{d}N(t) := \sum_i f(t_i)$.

Hence, $\mathcal{D}(\boldsymbol{u})$ denotes the difference of the log-likelihood between these two point processes:

$$\begin{aligned}
\mathcal{D}(\boldsymbol{u}) &= \mathcal{L}(\boldsymbol{\lambda}(t)) - \mathcal{L}(\tilde{\boldsymbol{\lambda}}(\boldsymbol{u}(t), t)) \\
&= \sum_{i=1}^{M} \bigg( \int_0^T \Big(\tilde{\lambda}_i(u_i(s), s) - \lambda_i(s)\Big)\mathrm{d}s - \int_0^T \log\Big(\frac{\tilde{\lambda}_i(u_i(s), s)}{\lambda_i(s)}\Big)\mathrm{d}N_i(s)\bigg) \\
&= \sum_{i=1}^{M} \bigg( \int_0^T \Big(u_i(s)\lambda_i(s) - \lambda_i(s)\Big)\mathrm{d}s - \int_0^T \log\Big(u_i(s)\Big)\mathrm{d}N_i(s)\bigg) \\
&= \sum_{i=1}^{M} \bigg( \int_0^T \Big(u_i(s) - 1\Big)\lambda_i(s)\mathrm{d}s - \int_0^T \log\Big(u_i(s)\Big)\mathrm{d}N_i(s)\bigg)
\end{aligned} \tag{30}$$

where $M$ is the dimension of point process. (30) comes from the form of control in (4). $\lambda_i(t), N_i(t), u_i(t)$ denote the $i$-th dimension of $\boldsymbol{\lambda}(t), \boldsymbol{N}(t), \boldsymbol{u}(t)$.

□

## C. Derivations of the Optimal Control Policy in (14)

We will formulate our objective function based on the form of optimal measure $\mathbb{Q}^*$ in (10). More specifically, we find a control $\boldsymbol{u}$ which pushes the controlled measure $\mathbb{Q}(\boldsymbol{u})$, as close to the optimal measure as possible. This leads to minimizing the Kullback-Leibler (KL) distance:

$$\boldsymbol{u}^* = \operatorname*{argmin}_{\boldsymbol{u}>0} \mathbb{D}_{KL}(\mathbb{Q}^* || \mathbb{Q}(\boldsymbol{u})) \tag{31}$$

This objective function is in sharp contrast to traditional methods that solve the optimal control problem by computing the solution the HJB PDE, which have severe limitations in scalability and feasibility to nonlinear jump diffusion SDEs.

Next we simplify the objective function. According to the definition of KL divergence and chain rule of derivatives, we have:

$$\mathbb{D}_{KL}(\mathbb{Q}^* || \mathbb{Q}(\boldsymbol{u})) = \mathbb{E}_{\mathbb{Q}^*}\left[\log\left(\frac{\mathrm{d}\mathbb{Q}^*}{\mathrm{d}\mathbb{Q}(\boldsymbol{u})}\right)\right] = \mathbb{E}_{\mathbb{Q}^*}\left[\log\left(\frac{\mathrm{d}\mathbb{Q}^*}{\mathrm{d}\mathbb{P}}\frac{\mathrm{d}\mathbb{P}}{\mathrm{d}\mathbb{Q}(\boldsymbol{u})}\right)\right] \tag{32}$$

The derivative $\mathrm{d}\mathbb{Q}^*/\mathrm{d}\mathbb{P}$ is given in (19) and $\mathrm{d}\mathbb{P}/\mathrm{d}\mathbb{Q}(\boldsymbol{u})$ is given in Theorem 3. Hence, we then substitute $\mathrm{d}\mathbb{Q}^*/\mathrm{d}\mathbb{P}$ and $\mathrm{d}\mathbb{P}/\mathrm{d}\mathbb{Q}(\boldsymbol{u})$ to (32). After removing terms which are independent of $\boldsymbol{u}$, the objective function (31) is simplified as:

$$\boldsymbol{u}^* = \operatorname*{argmin}_{\boldsymbol{u}>0} \mathbb{E}_{\mathbb{Q}^*}[\mathcal{D}(\boldsymbol{u})]$$

Next we parameterize $\boldsymbol{u}(t)$ as a piecewise constant function on $[0, T]$ as follows.

$$\boldsymbol{u}(t) = \begin{cases} \vdots \\ \boldsymbol{u}^k & \text{for } t \in [k\Delta t, (k+1)\Delta t) \\ \vdots \end{cases}$$

More specifically, the $k$-th piece is defined on $[k\Delta t, (k+1)\Delta t)$ as $\boldsymbol{u}^k$, where $k = 0, \cdots, K-1$, $t_k = k\Delta t$ and $T = t_K$. Then we have:

$$\mathbb{E}_{\mathbb{Q}^*}[\mathcal{D}(\boldsymbol{u})] = \sum_{i=1}^{M}\sum_{k=1}^{K}\left(\mathbb{E}_{\mathbb{Q}^*}\Big[\int_{t_k}^{t_{k+1}} (u_i^k - 1)\lambda_i(s)\mathrm{d}s\Big] - \mathbb{E}_{\mathbb{Q}^*}\Big[\int_{t_k}^{t_{k+1}} \log(u_i^k)\mathrm{d}N_i(s)\Big]\right) \tag{33}$$

where $u_i^k$ is the $i$-th dimension of $\boldsymbol{u}^k$. To compute $u_i^k$, we can neglect the two summation terms in (33) and only focus on the parts that involves $u_i^k$. Then we move $u_i^k$ outside of the expectation and discard any constant terms. This yields the function that only involves $u_i^k$:

$$f(u_i^k) = u_i^k \mathbb{E}_{\mathbb{Q}^*}\Big[\int_{t_k}^{t_{k+1}} \lambda_i(s)\mathrm{d}s\Big] - \log(u_i^k)\mathbb{E}_{\mathbb{Q}^*}\Big[\int_{t_k}^{t_{k+1}} \mathrm{d}N_i(s)\Big] \tag{34}$$

We can then show $f(u_i^k)$ is convex in $u_i^k$. More specifically, it is in the form of $f(x) = ax - \log(x)b$ with $a > 0, b > 0$ and $f''(x) > 0$. Finally, setting $f'(u_i^k) = 0$ yields $u_i^{k*}$:

$$u_i^{k*} = \frac{\mathbb{E}_{\mathbb{Q}^*}\big[\int_{t_k}^{t_{k+1}} \mathrm{d}N_i(s)\big]}{\mathbb{E}_{\mathbb{Q}^*}\big[\int_{t_k}^{t_{k+1}} \lambda_i(s)\mathrm{d}s\big]} \tag{35}$$

However, $u_i^{k*}$ is still not computable since the expectation is taken under the optimal probability measure $\mathbb{Q}^*$. Since we only known the SDE of the uncontrolled dynamics and can only compute the expectation under $\mathbb{P}$, we need to change the expectation from $\mathbb{E}_{\mathbb{Q}^*}$ to $\mathbb{E}_{\mathbb{P}}$ to compute $u_i^{k*}$.

To do this, we will use the following lemma.

**Lemma 4.** *Let the probability measure* $\mathbb{Q}^*$ *be defined as* $\frac{\mathrm{d}\mathbb{Q}^*}{\mathrm{d}\mathbb{P}} = \frac{\exp(-\frac{1}{\gamma}S(\boldsymbol{x}))}{\mathbb{E}_{\mathbb{P}}[\exp(-\frac{1}{\gamma}S(\boldsymbol{x}))]}$ *in (10), and* $g(\boldsymbol{x}) : \Omega \to \Re$ *be any measurable function. Then we have:*

$$\mathbb{E}_{\mathbb{Q}^*}[g(\boldsymbol{x})] = \frac{\mathbb{E}_{\mathbb{P}}\Big[\exp\big(-\frac{1}{\gamma}S(\boldsymbol{x})\big)g(\boldsymbol{x})\Big]}{\mathbb{E}_{\mathbb{P}}[\exp(-\frac{1}{\gamma}S(\boldsymbol{x}))]}$$

*Proof.*

$$\begin{aligned}
\mathbb{E}_{\mathbb{Q}^*}[g(\boldsymbol{x})] &= \int g(\boldsymbol{x})\mathrm{d}\mathbb{Q}^* \\
&= \int g(\boldsymbol{x})\frac{\exp(-\frac{1}{\gamma}S(\boldsymbol{x}))\mathrm{d}\mathbb{P}}{\mathbb{E}_{\mathbb{P}}[\exp(-\frac{1}{\gamma}S(\boldsymbol{x}))]} \\
&= \frac{\int \Big(g(\boldsymbol{x})\exp\big(-\frac{1}{\gamma}S(\boldsymbol{x})\big)\Big)\mathrm{d}\mathbb{P}}{\mathbb{E}_{\mathbb{P}}[\exp(-\frac{1}{\gamma}S(\boldsymbol{x}))]} \\
&= \frac{\mathbb{E}_{\mathbb{P}}\Big[\exp\big(-\frac{1}{\gamma}S(\boldsymbol{x})\big)g(\boldsymbol{x})\Big]}{\mathbb{E}_{\mathbb{P}}[\exp(-\frac{1}{\gamma}S(\boldsymbol{x}))]}
\end{aligned}$$

□

Finally, applying Lemma 4 to (35) yields the following expression for the optimal policy:

$$u_i^{k*} = \frac{\mathbb{E}_{\mathbb{Q}^*}\big[\int_{t_k}^{t_{k+1}}\mathrm{d}N_i(s)\big]}{\mathbb{E}_{\mathbb{Q}^*}\big[\int_{t_k}^{t_{k+1}}\lambda_i(s)\mathrm{d}s\big]} = \frac{\frac{\mathbb{E}_{\mathbb{P}}\big[\exp(-\frac{1}{\gamma}S(\boldsymbol{x}))\int_{t_k}^{t_{k+1}}\mathrm{d}N_i(s)\big]}{\mathbb{E}_{\mathbb{P}}\big[\exp(-\frac{1}{\gamma}S(\boldsymbol{x}))\big]}}{\frac{\mathbb{E}_{\mathbb{P}}\big[\exp(-\frac{1}{\gamma}S(\boldsymbol{x}))\int_{t_k}^{t_{k+1}}\lambda_i(s)\mathrm{d}s\big]}{\mathbb{E}_{\mathbb{P}}\big[\exp(-\frac{1}{\gamma}S(\boldsymbol{x}))\big]}} = \frac{\mathbb{E}_{\mathbb{P}}\big[\exp(-\frac{1}{\gamma}S(\boldsymbol{x}))\int_{t_k}^{t_{k+1}}\mathrm{d}N_i(s)\big]}{\mathbb{E}_{\mathbb{P}}\big[\exp(-\frac{1}{\gamma}S(\boldsymbol{x}))\int_{t_k}^{t_{k+1}}\lambda_i(s)\mathrm{d}s\big]} \tag{36}$$

The derivation of the optimal policy is now complete.

## D. Derivations of the Control Cost

We will derive the control cost in (9), which comes naturally from the dynamics. According to the definition of the KL divergence, we have:

$$\mathbb{D}_{KL}(\mathbb{Q}||\mathbb{P}) := \mathbb{E}_{\mathbb{Q}}[\log(\frac{\mathrm{d}\mathbb{Q}}{\mathrm{d}\mathbb{P}})] = \mathbb{E}_{\mathbb{Q}}[C(\boldsymbol{u})] \tag{37}$$

Hence, the next step is to compute the derivative $\frac{\mathrm{d}\mathbb{Q}}{\mathrm{d}\mathbb{P}}$. This derivative means the relative density of probability distribution $\mathbb{Q}$ with respect to $\mathbb{P}$. According to (Brémaud, 1981), we have:

$$\frac{\mathrm{d}\mathbb{Q}}{\mathrm{d}\mathbb{P}} = \exp\Big(\sum_i \int_0^T \log\Big(\frac{\tilde{\lambda}_i(u_i(t),t)}{\lambda_i(t)}\Big)\mathrm{d}N_i(u_i(t),t) - \int_0^T (\tilde{\lambda}_i(u_i(t),t) - \lambda_i(t))\mathrm{d}t\Big), \tag{38}$$

Using the relationship that $\lambda_i(u_i(t),t) = \lambda_i(t)u_i(t)$, we have:

$$\mathbb{E}_{\mathbb{Q}}[\log(\frac{\mathrm{d}\mathbb{Q}}{\mathrm{d}\mathbb{P}})] = \mathbb{E}_{\mathbb{Q}}\Big[\sum_i \int_0^T \log\Big(\frac{\tilde{\lambda}_i(u_i(t),t)}{\lambda_i(t)}\Big)\mathrm{d}\tilde{N}_i(u_i(t),t) - \int_0^T (\tilde{\lambda}_i(u_i(t),t) - \lambda_i(t))\mathrm{d}t\Big] \tag{39}$$

$$= \mathbb{E}_{\mathbb{Q}}\Big[\sum_i \int_0^T \log\Big(u_i(t)\Big)\mathrm{d}\tilde{N}_i(u_i(t),t) - \int_0^T \Big(1 - \frac{1}{u_i(t)}\Big)\tilde{\lambda}_i(u_i(t),t)\mathrm{d}t\Big] \tag{40}$$

$$= \mathbb{E}_{\mathbb{Q}}\Big[\sum_i \int_0^T \log\Big(u_i(t)\Big)\tilde{\lambda}_i(u_i(t),t)\mathrm{d}t + \int_0^T \Big(1 - \frac{1}{u_i(t)}\Big)\tilde{\lambda}_i(u_i(t),t)\mathrm{d}t\Big] \tag{41}$$

Note that (40) to (41) follows from the Campbell theorem (Daley & Vere-Jones, 2007). Therefore, the control cost is:

$$\begin{aligned} C(\boldsymbol{u}) &= \int_0^T \sum\nolimits_i \Big(\log(u_i(t)) + \frac{1}{u_i(t)} - 1\Big)\tilde{\lambda}_i(u_i(t),t)\mathrm{d}t \\ &= \int_0^T \sum\nolimits_i \Big(\log(u_i(t)) + \frac{1}{u_i(t)} - 1\Big)u_i(t)\lambda_i(t)\mathrm{d}t \end{aligned}$$

## E. Applying Our Framework to Traditional Stochastic Optimal Control Problem

In the traditional stochastic optimal control problem, the control policy $\boldsymbol{u}(t)$ is affine in the drift as follows:

$$\mathrm{d}\boldsymbol{x}(t) = \underbrace{\big(\boldsymbol{f}(\boldsymbol{x}) + \boldsymbol{G}(\boldsymbol{x})\boldsymbol{u}(t)\big)\mathrm{d}t}_{\text{drift}} + \underbrace{\boldsymbol{g}(\boldsymbol{x})\mathrm{d}\boldsymbol{w}(t)}_{\text{diffusion}} + \underbrace{\boldsymbol{h}(\boldsymbol{x})\mathrm{d}\boldsymbol{N}(t)}_{\text{jump}} \tag{42}$$

where $\boldsymbol{G}(\boldsymbol{x}) : \Re^K \rightarrow \Re^K$. Note that in this case, the jump process $\boldsymbol{N}(t)$ is not controlled. The goal is also to find $\boldsymbol{u}^* \in \Re^K$ such that

$$\boldsymbol{u}^* = \operatorname*{argmin}_{\boldsymbol{u}} \mathbb{E}_{\mathbb{Q}}[S(\boldsymbol{x})] + \gamma * \text{control cost}, \tag{43}$$

Next we use our framework to solve the problem as follows. First, the objective is the same as (11):

$$\boldsymbol{u}^* = \operatorname*{argmin}_{\boldsymbol{u}\in\mathcal{U}} \mathbb{D}_{KL}\big(\mathbb{Q}^* || \mathbb{Q}(\boldsymbol{u})\big) \tag{44}$$

where $\mathcal{U}$ is the set of admissible control policies. Next we will also simplify the objective function. According to the definition of relative entropy and chain rule of derivatives, we have:

$$\mathbb{D}_{KL}(\mathbb{Q}^* || \mathbb{Q}(\boldsymbol{u})) = \mathbb{E}_{\mathbb{Q}^*}\left[\log\left(\frac{\mathrm{d}\mathbb{Q}^*}{\mathrm{d}\mathbb{P}}\frac{\mathrm{d}\mathbb{P}}{\mathrm{d}\mathbb{Q}(\boldsymbol{u})}\right)\right] \tag{45}$$

The derivative $\mathrm{d}\mathbb{Q}^*/\mathrm{d}\mathbb{P}$ is given in (10). Hence we just need to compute $\mathrm{d}\mathbb{P}/\mathrm{d}\mathbb{Q}(\boldsymbol{u})$. Intuitively, the derivative $\mathrm{d}\mathbb{P}/\mathrm{d}\mathbb{Q}(\boldsymbol{u})$ means the relative density of probability distribution $\mathbb{P}$ with respect to $\mathbb{Q}$.

In the intensity control problem, the change of measure happens because the intensity of the temporal point process that drives the SDE in (2) is changed from $\boldsymbol{\lambda}(t)$ to $\boldsymbol{\lambda}(\boldsymbol{u}, t)$. However, in the traditional problem, the change of measure is due to the fact that the drift term in the SDE is changed by the control policy.

Hence according to the classic Girsanov's theorem (Hanson, 2007) in probability theory, it is in the form:

$$\frac{\mathrm{d}\mathbb{P}}{\mathrm{d}\mathbb{Q}(\boldsymbol{u})} = \exp\big(\mathcal{D}(\boldsymbol{u})\big),$$

where $\mathcal{D}(\boldsymbol{u})$ is defined as:

$$\mathcal{D}(\boldsymbol{u}) = -\int_0^T \boldsymbol{u}(t)^\top \boldsymbol{G}(\boldsymbol{x})^\top \boldsymbol{\Sigma}^{-1}\boldsymbol{g}(\boldsymbol{x})\mathrm{d}\boldsymbol{w} + \frac{1}{2}\int_0^T \boldsymbol{u}(t)^\top \boldsymbol{G}(\boldsymbol{x})^\top \boldsymbol{\Sigma}^{-1}\boldsymbol{u}(t)\mathrm{d}t$$

where $\boldsymbol{\Sigma} = \boldsymbol{g}\boldsymbol{g}^\top$. We then substitute $\mathrm{d}\mathbb{Q}^*/\mathrm{d}\mathbb{P}$ and $\mathrm{d}\mathbb{P}/\mathrm{d}\mathbb{Q}(\boldsymbol{u})$ to (45). After removing terms which are independent of $\boldsymbol{u}$, (44) is simplified as:

$$\boldsymbol{u}^* = \operatorname*{argmin}_{\boldsymbol{u}\in\mathcal{U}} \mathbb{E}_{\mathbb{Q}^*}[\mathcal{D}(\boldsymbol{u})] \tag{46}$$

Since we apply the control in discrete timestamps, it suffices to consider $\mathcal{U}$ as the class of piecewise constant functions on $[0, T]$, with the $k$-th piece of $\boldsymbol{u}$ defined on $[k\Delta t, (k+1)\Delta t)$ as $\boldsymbol{u}^k$, where $k = 0, \cdots, K-1$, $t_k = k\Delta t$ and $T = t_K$. Then we express the objective function as:

$$\mathbb{E}_{\mathbb{Q}^*}[\mathcal{D}(\boldsymbol{u})] = \sum_{k=1}^{K}\left(\frac{1}{2}\boldsymbol{u}^{k\top}\mathbb{E}_{\mathbb{Q}^*}\big[\int_{t_k}^{t_{k+1}} \boldsymbol{G}^\top\boldsymbol{\Sigma}^{-1}\boldsymbol{G}\mathrm{d}t\big]\boldsymbol{u}^k - \boldsymbol{u}^{k\top}\mathbb{E}_{\mathbb{Q}^*}\big[\int_{t_k}^{t_{k+1}} \boldsymbol{G}^\top\boldsymbol{\Sigma}^{-1}\boldsymbol{g}\mathrm{d}\boldsymbol{w}\big]\right) \tag{47}$$

We can neglect the summation term in (47) and only focus on the parts that involves $\boldsymbol{u}^k$. Since the expression is quadratic in $\boldsymbol{u}^k$, it is convex in $\boldsymbol{u}^k$. Finally, setting the gradient to zero yields:

$$\boldsymbol{u}^{k*} = \frac{\mathbb{E}_{\mathbb{Q}^*}\big[\int_{t_k}^{t_{k+1}} \boldsymbol{G}^\top\boldsymbol{\Sigma}^{-1}\boldsymbol{g}\mathrm{d}\boldsymbol{w}\big]}{\mathbb{E}_{\mathbb{Q}^*}\big[\int_{t_k}^{t_{k+1}} \boldsymbol{G}^\top\boldsymbol{\Sigma}^{-1}\boldsymbol{G}\mathrm{d}t\big]} = \frac{\mathbb{E}_{\mathbb{P}}\big[\exp(-\frac{1}{\gamma}S(\boldsymbol{x}))\int_{t_k}^{t_{k+1}} \boldsymbol{G}^\top\boldsymbol{\Sigma}^{-1}\boldsymbol{g}\mathrm{d}\boldsymbol{w}\big]}{\mathbb{E}_{\mathbb{P}}\big[\exp(-\frac{1}{\gamma}S(\boldsymbol{x}))\int_{t_k}^{t_{k+1}} \boldsymbol{G}^\top\boldsymbol{\Sigma}^{-1}\boldsymbol{G}\mathrm{d}t\big]} \tag{48}$$

Then we can also use Algorithm 1 to compute the optimal control policy.